\documentclass[runningheads]{llncs}

\makeatletter
\def\thanks#1{\protected@xdef\@thanks{\@thanks
        \protect\footnotetext{#1}}}
\makeatother

\usepackage{eccv}

\usepackage{eccvabbrv}

\usepackage{graphicx}
\usepackage{booktabs}
\usepackage{amssymb}
\usepackage{pifont}
\usepackage{subcaption}
\usepackage[accsupp]{axessibility}  

\usepackage{hyperref}

\usepackage{orcidlink}

\usepackage{pifont}       
\usepackage{bbding}       
\usepackage{fontawesome}  

\begin{document}


\title{Beyond MOT: Semantic Multi-Object Tracking} 

\titlerunning{Beyond MOT: Semantic Multi-Object Tracking}

\author{
Yunhao Li\inst{1,2} \and
Qin Li\inst{1} \and
Hao Wang\inst{2} \and
Xue Ma\inst{1} \and
Jiali Yao\inst{3} \and
Shaohua Dong\inst{4} \and \\
Heng Fan\inst{4\dag} \and
Libo Zhang\inst{1,2,3\dag\sharp}
\thanks{$^{\dag}$Equal advising and co-last author \; $^{\sharp}$Corresponding author (libo@iscas.ac.cn)}
}

\authorrunning{Y.~Li et al.}

\institute{Institute of Software Chinese Academy of Sciences \and
University of Chinese Academy of Sciences \and
Hangzhou Institute for Advanced Study, University of Chinese Academy of Sciences \and
Department of Computer Science \& Engineering, University of North Texas\\
}

\maketitle

\begin{abstract}


Current multi-object tracking (MOT) aims to predict trajectories of targets (\ie, ``\emph{where}'') in videos. Yet, knowing merely ``\emph{where}'' is insufficient in many crucial applications. In comparison, semantic understanding such as fine-grained behaviors, interactions, and overall summarized captions (\ie, ``\emph{what}'') from videos, associated with ``\emph{where}'', is highly-desired for comprehensive video analysis. Thus motivated, we introduce \emph{Semantic Multi-Object Tracking} (SMOT), that aims to estimate object trajectories and meanwhile understand semantic details of associated trajectories including instance captions, instance interactions, and overall video captions, integrating ``\emph{where}'' and ``\emph{what}'' for tracking. In order to foster the exploration of SMOT, we propose BenSMOT, a large-scale Benchmark for Semantic MOT. Specifically, BenSMOT comprises 3,292 videos with 151K frames, covering various scenarios for semantic tracking of humans. BenSMOT provides annotations for the trajectories of targets, along with associated instance captions in natural language, instance interactions, and overall caption for each video sequence. To our best knowledge, BenSMOT is the \emph{first} publicly available benchmark for SMOT. Besides, to encourage future research, we present a novel tracker named SMOTer, which is specially designed and end-to-end trained for SMOT, showing promising performance. By releasing BenSMOT, we expect to go beyond conventional MOT by predicting ``\emph{where}'' and ``\emph{what}'' for SMOT, opening up a new direction in tracking for video understanding. We will release BenSMOT and SMOTer at \url{https://github.com/Nathan-Li123/SMOTer}.

\end{abstract}

\section{Introduction}
\label{sec:intro}

Multi-object tracking (MOT) is a fundamental problem in computer vision with many applications such as autonomous driving, robotics, and visual surveillance. Current MOT tasks (\eg,~\cite{milan2016mot16,geiger2013vision,dave2020tao,yu2020bdd100k,sun2022dancetrack}) usually focus on predicting trajectories of the targets from videos, \ie, answering the question of ``\emph{where are the targets}''. Despite considerable advancements in deep learning era (\eg,~\cite{yan2022towards,meinhardt2022trackformer,zhou2022global,zhang2023motrv2,gao2023memotr}), knowing merely ``\emph{where}'' in existing MOT is \emph{insufficient} for video understanding in many important applications. For example, besides trajectories, knowing more trajectory-associated \emph{semantic} details of the objects, such as the behaviors and their interactions with the surroundings (\ie, ``\emph{what}''), is greatly beneficial for visual surveillance and robotics, which grows the interest in expanding modern MOT (\ie, addressing ``\emph{where}'') with trajectory-associated semantic analysis (\ie, understanding ``\emph{what}'') for more comprehensive video understanding.

\begin{figure*}[!t]
 \centering
 \includegraphics[width=0.95\linewidth]{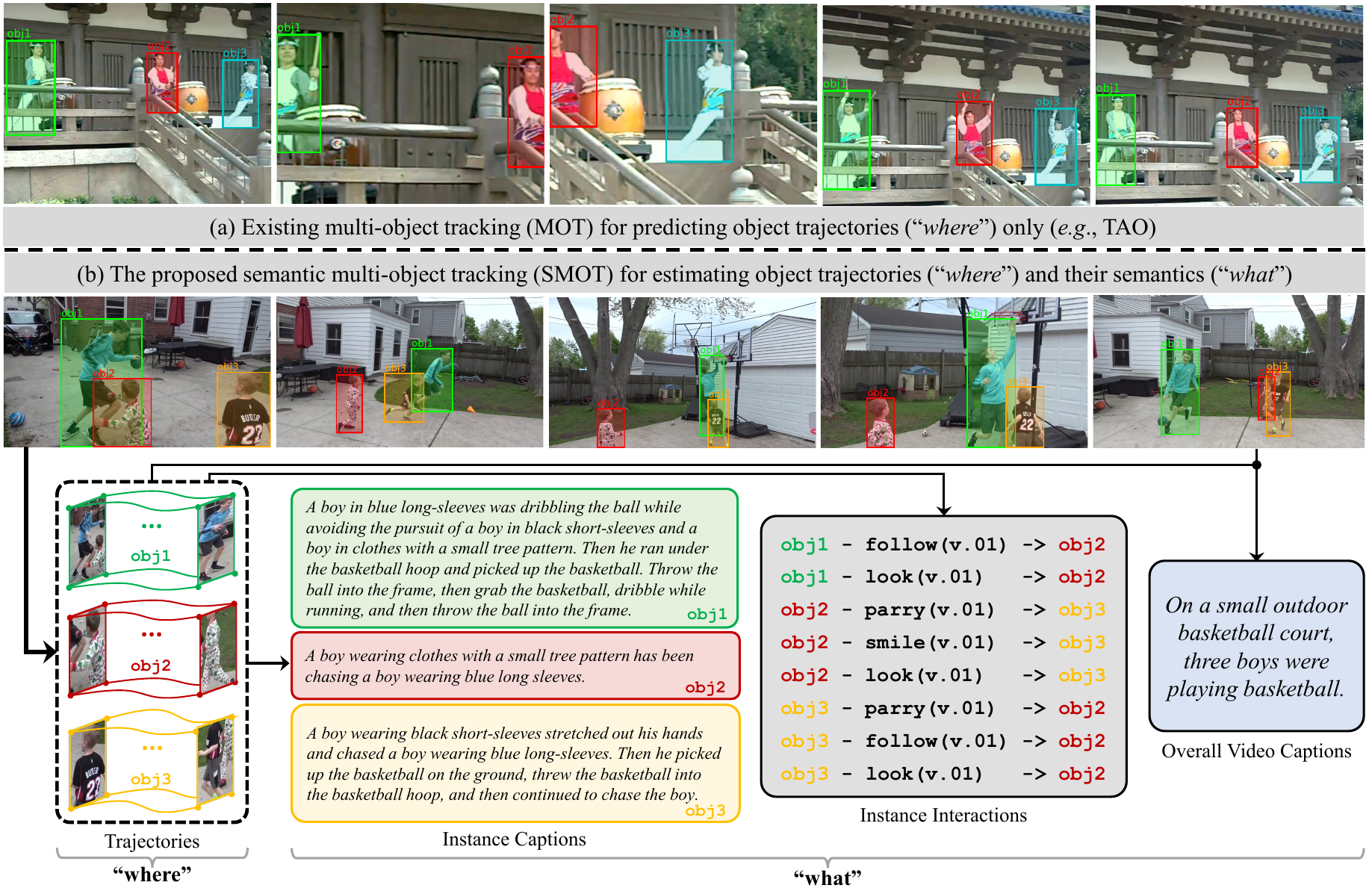}
 \caption{Existing multi-object tracking (MOT) focusing on predicting trajectories only (see (a)) and our semantic multi-object tracking (SMOT) aiming at estimating trajectories and understanding their semantics (see (b)). Best viewed in color for all figures.}
 \label{fig:smot}
\end{figure*}

Thus motivated, in this work we introduce a new type of MOT task, dubbed \emph{Semantic Multi-Object Tracking} (\textbf{SMOT})\footnote{Here by “semantic”, we emphasize high-level trajectory-based activity understanding in videos in the context of tracking, instead of category as in semantic segmentation.}, which aims to expand scope of MOT beyond merely predicting trajectories to capture rich semantic details of objects in videos. Specifically, besides trajectory estimation, SMOT comprises three additional trajectory-associated semantic understanding tasks, including \emph{instance captioning}, \emph{instance interaction recognition}, and \emph{video captioning}, as illustrated in Fig.~\ref{fig:smot}. In particular, instance captioning aims at describing objects and their behaviors in human language, answering ``\emph{what are the objects doing}''; instance interaction recognition is to capture relations between objects, answering ``\emph{what are the relations between objects}''; video captioning provides the overall scenario understanding based on trajectories, answering ``\emph{what is happening}''. Note that, all these three additional tasks are associated with object trajectories in videos. Ultimately, the goal of SMOT is to integrate ``\emph{where}'' (\ie, object trajectories) and ``\emph{what}'' (\ie, trajectory-associated semantics) for video understanding, going beyond MOT predicting only trajectories (see again Fig.~\ref{fig:smot}). Compared to conventional MOT, our SMOT is a multimodal task across vision and language, which is more challenging yet practical. It is worthy to notice that, SMOT is a natural extension of MOT. Despite sharing some similarities with video captioning~\cite{krishna2017dense}, SMOT \emph{differs} in that it aims at language understanding of dense target trajectories from videos in the context of instance-level tracking.


To foster study of SMOT, we propose \textbf{BenSMOT}, a large-scale Benchmark for Semantic MOT. Specifically, BenSMOT contains 3,292 video sequences with 151K frames, captured from more than 40 diverse daily-life scenarios for \emph{\textbf{human-centric}} semantic tracking. In BenSMOT, there exist more than 7.8K instances, labeled with over 335K bounding boxes for trajectory prediction. For each instance, we provide fine-grained human language to describe the behaviors for instance captioning, resulting in 7.8K sentences. In addition, to capture interactions between objects, BenSMOT provides more than 14K interaction annotations from a rich collection of 335 kinds of interactions. Moreover, for each sequence, a summary text is provided for overall understanding of instances and backgrounds, leading to 3,292 video caption texts in total in BenSMOT. To ensure the high quality of BenSMOT, annotations are manually labeled with careful inspection and refinement. \emph{To our best knowledge}, BenSMOT is the \textbf{first} publicly available benchmark dedicated to SMOT. By releasing BenSMOT, we expect it to serve as a platform for advancing the research on SMOT.

Furthermore, in order to facilitate the development of SMOT algorithms on BenSMOT, we present a simple yet effective tracker dubbed \textbf{SMOTer}. Specifically, SMOTer is built upon a multi-object tracking model~\cite{zhang2022bytetrack}. After generating object trajectories, we extract features for each trajectory and design additional prediction heads for semantic understanding of instance captioning, interaction recognition, and video captioning. Note, SMOTer is \emph{not} a simple offline assembly of models for different tasks. Instead, it is specially designed for SMOT and trained in an end-to-end manner for predicting trajectories and understanding their semantics in videos. Despite simplicity, SMOTer shows promising performance for SMOT, and outperforms the offline-combination strategies, evidencing its effectiveness. We expect it to provide a reference for future research.

We notice there exists a concurrent work combining multi-object tracking and trajectory captioning for dense video object captioning~\cite{zhou2023dense}. Compared with~\cite{zhou2023dense}, our work differs in three aspects. First, \emph{task-wise}, besides trajectory estimation and captioning, the SMOT in this work contains instance interaction and overall video captioning, providing more semantic details. Second, \emph{dataset-wise}, we introduce a large-scale benchmark, BenSMOT, that is dedicated for SMOT and supports complete end-to-end model training, while the work of~\cite{zhou2023dense} borrows datasets from different tasks, which results in disjoint training of the model on different tasks and may thus degrade performance. Third, \emph{model-wise}, owing to BenSMOT, we propose an end-to-end algorithm that demonstrates promising results on SMOT, while the approach in~\cite{zhou2023dense} is disjointly learned and does not support predicting instance interactions and overall video captioning.

In summary, we make the following contributions: \ding{171} We introduce semantic multi-object tracking (SMOT), a new tracking paradigm that expands existing MOT task by integrating ``\emph{where}'' and ``\emph{what}''; \ding{170} We present BenSMOT, a large-scale dataset with 3,292 videos and rich annotations for SMOT; \ding{168} We propose SMOTer, a simple but effective tracker to facilitate future research of SMOT; \ding{169} We show that SMOTer achieves promising performance for SMOT and conduct in-depth analysis on it to provide guidance for future algorithm design.

\section{Related Works}
\label{relatedworks}

\noindent
\textbf{MOT Benchmarks.} Benchmarks have played a crucial role in facilitating the development of MOT. PETS2009~\cite{ferryman2009pets2009} is one of the earliest benchmarks for multi-pedestrian tracking. Later, the popular MOT Challenge~\cite{leal2015motchallenge, milan2016mot16, dendorfer2020mot20} has been introduced with more crowded videos  and greatly advanced MOT. KITTI~\cite{geiger2013vision} and BDD100K~\cite{yu2020bdd100k} are specifically designed for object tracking in autonomous driving. ImageNet-Vid~\cite{deng2009imagenet} provides trajectory annotations for 30 categories across over 1,000 videos, and TAO~\cite{dave2020tao} further expands object classes to 833 for generic multi-object tracking. To foster MOT in specific dancing and sport scenarios, DanceTrack~\cite{sun2022dancetrack} and SportsMOT~\cite{cui2023sportsmot} have been presented for dancer and player tracking.  AnimalTrack~\cite{zhang2023animaltrack} focuses on tracking various animals in wild scenes. Moreover, UAVDT~\cite{du2018unmanned} and VisDrone~\cite{zhu2021detection} provide platforms for tracking targets with drones. \textbf{\emph{Different from}} the above MOT benchmarks for object trajectory prediction only (``\emph{where}''), BenSMOT is specially developed for the new task of SMOT to serve as a platform for simultaneous estimation of target trajectories and trajectory-associated semantic understanding. To this end, it offers rich annotations of not only object trajectories as in existing standard MOT datasets (``\emph{where}'') but also semantic details of objects (``\emph{what}'') such as instance trajectory captions, interactions and overall video captions (see Fig.~\ref{fig:smot} again).

\noindent
\textbf{MOT Algorithms.} MOT algorithms have seen great progress in recent years. One of the popular paradigms for multi-object tracking is the so-called \emph{Tracking-by-Detection}. This paradigm involves initial object detection followed by association based on these detections, forming the basis for many representative methods~\cite{wojke2017simple, he2016deep, du2023strongsort, cao2023observation, maggiolino2023deep, zhang2022bytetrack}. In this case, MOT methods typically improve their performance by enhancing both detection and matching effectiveness. In addition, another prevalent paradigm is the \emph{Joint-Tracking-and-Detection}~\cite{ye2022joint, yan2022towards, zhou2020tracking, han2020unicorn, yan2023universal}, which integrates tracking and detection steps into a single process, enabling end-to-end training. Recently, Transformer~\cite{vaswani2017attention} has been introduced into MOT and exhibited remarkable improvements over previous tackers~\cite{sun2020transtrack, meinhardt2022trackformer, zeng2022motr, chu2023transmot, zhang2023motrv2, gao2023memotr, zhou2022global}. Our SMOTer is related to but \emph{\textbf{different than}} existing MOT methods. Specifically, going beyond merely predicting object trajectories, SMOTer also aims at semantic understanding of trajectories, integrating ``\emph{where}'' and ``\emph{what}''.

\noindent
\textbf{Video Captioning.} 
Video captioning is a multimodal vision-language task that aims to automatically describe the video content using natural language. Owing to its important applications in video event commentary and human-computer interaction, video captioning has drawn increasing attention in the past decade with many models proposed (\eg,~\cite{krishna2017dense,zhou2018end,seo2022end,lin2022swinbert,yang2023vid2seq,shen2023accurate}). Similar to the video captioning task, SMOT utilizes natural language to describe instances and the overall video content. However, the \emph{\textbf{difference}} is that our SMOT aims at language comprehension for dense target trajectories in the context of multi-object tracking, which is more challenging due to the requirement of accurate trajectories yet provides more fine-grained information for video understanding.

\noindent
\textbf{Visual Relationship Detection.} The goal of visual relationship detection task is to identify and comprehend the relationships (usually represented by a triplet of <\emph{subject}, \emph{predicate}, \emph{object}>) between objects from an image (\eg,~\cite{lu2016visual,liang2017deep,zhang2017visual}). Besides static images, recent works extend visual relationship detection in the video domain (\eg,~\cite{liu2020beyond,chen2021social,zheng2022vrdformer}). Our work is relevant to video visual relationship detection but \emph{\textbf{different}} by extending this task target-specific trajectories, and thus more challenging due to complicated multi-object tracking scenarios.

\section{The Proposed BenSMOT}

\subsection{Design Principle}

BenSMOT aims at providing a new platform for exploring human-centric SMOT. In construction of BenSMOT, we follow the following rules: 

(1) \emph{Dedicated benchmark}. The key motivation of our BenSMOT is to provide a dedicated dataset for semantic multi-object tracking. In current deep learning era, a large number of videos are desired in benchmark construction for training robust tracking models. Considering this, we hope to build a dedicated dataset with more than 3,000 videos for human-centric semantic multi-object tracking. 

(2) \emph{Diverse scenarios}. For a dataset, diversity is crucial, in both training and evaluation, for developing general systems. In order to provide a diverse platform for SMOT, in BenSMOT we will include videos from more than 40 various scenes that range from daily life scenarios to dancing scenarios to sport scenarios. 

(3) \emph{High-quality annotations}. High-quality annotation is crucial for a benchmark in both training and assessing models. In BenSMOT, we ensure its high quality by providing manual annotations for each sequence including object trajectories, instance captions, instance interactions and overall video captions. In this process, multi-round inspections and refinements will be carried out.

\subsection{Data Acquisition}
BenSMOT focuses on predicting the target trajectories and meanwhile understanding their semantics by instance captioning, interaction recognition, and overall video captioning. To this end, the video sequences in BenSMOT are desired to involve diverse multi-person activities and interactions, which is different from videos in existing benchmarks to some extent. Specifically, we first identify 47 common scenarios, involving different activities ranging from daily-life talk and play to different sports, that are suitable for tracking by drawing inspiration from~\cite{DBLP:conf/cvpr/HeilbronEGN15}. Due to limited space, please refer to \textbf{supplementary material} for detailed scenarios in BenSMOT. Afterwards, we search for raw sequences for each scenario under the {\tt Creative Commons License} from YouTube, the largest and most popular video platform with massive real-world videos. Please notice, for each raw video, instead of using the whole sequence, we usually choose a suitable clip for our semantic multi-object tracking task. 

\begin{table*}[t]
\centering
\renewcommand{\arraystretch}{1}
\tabcolsep=0.8mm
\caption{Summary of BenSMOT and its comparison with popular multi-object tracking benchmarks. "n/a" denotes that data is not available.}
\resizebox{0.9\textwidth}{!}{
\begin{tabular}{l|ccccccccc}
    \toprule[1.2pt]
    & KITTI & MOT17 & MOT20 & BDD100k & TAO & GMOT-40 & DanceTrack & SportsMOT & BenSMOT\\
    & ~\cite{geiger2013vision} & ~\cite{milan2016mot16} & ~\cite{dendorfer2020mot20} & ~\cite{yu2020bdd100k} & ~\cite{dave2020tao} & ~\cite{bai2021gmot} & ~\cite{sun2022dancetrack} & ~\cite{cui2023sportsmot} & (ours)\\ 
    \midrule[0.8pt]
    Videos & 50 & 14 & 8 & 1,600 & 2,907 & 40 & 100 & 240 & 3,292 \\
    Min. length (s) & n/a & 17.0 & 17.0 & 40.0 & n/a & 3.0 & n/a & n/a & 1.5 \\
    Avg. length (s) & 10.0 & 33.0 & 66.8 & 40.0 & 36.8 & 8.9 & 52.9 & n/a & 22.9 \\
    Max. length (s) & n/a & 85.0 & 133.0 & 40.0 & n/a & 24.2 & n/a & n/a & 116.0 \\
    Total length (s) & 498 & 463 & 535 & 640,000 & 106,978 & 356 & 5,292 & 6015 & 75,499 \\
    Total tracks & 2,600 & 1.3K & 3.83K & 131K & 17,287 & 2,026 & 990 & 3,401 & 7,792\\
    Total boxes & 80K & 300K & 2,102K & 3,300K & 333K & 256K & n/a & 1,629K & 335K\\
    Total frames & 15K & 11K & 13K & 318K & 2,674K & 9K & 106K & 150K & 151K\\
    Instance Captions & \ding{56} & \ding{56} & \ding{56} & \ding{56} & \ding{56} & \ding{56} & \ding{56} & \ding{56} & 7,792 \\
    Instance Interactions & \ding{56} & \ding{56} & \ding{56} & \ding{56} & \ding{56} & \ding{56} & \ding{56} & \ding{56} & 14K \\
    Video Summaries & \ding{56} & \ding{56} & \ding{56} & \ding{56} & \ding{56} & \ding{56} & \ding{56} & \ding{56} & 3,292 \\
    \bottomrule[1.2pt]
\end{tabular}
\label{tab:summary}
}	
\end{table*}

Eventually, we compile a dataset for SMOT by including 3,292 videos with 151K frames. BenSMOT has an average video length of 23 seconds, with our current focus in this work not being on long-term SMOT. The total number of instance tracks is 7.8K, annotated with 335K bounding boxes for tracking. Besides bounding boxes, we provide 7.8K instance captions, 14K interactions, and 3,292 video summaries for trajectory-associated semantic understanding (how to annotate will be described in the following). To our best knowledge, BenSMOT by far is the first publicly available benchmark dedicated for SMOT. Tab.~\ref{tab:summary} summarizes our BenSMOT and compares it with popular MOT benchmarks.


\subsection{Data Annotation}

In order to meet the requirements of SMOT, BenSMOT provides four types of annotations, including bounding box, instance caption, instance interaction, and the overall video caption. Specifically, for each trajectory in videos, similar to standard MOT benchmarks (\eg,~\cite{dave2020tao,milan2016mot16}), we provide axis-aligned bounding boxes to indicate its spatial positions in the videos. For instance captioning, we label each trajectory with a precise sentence in natural language to describe the detailed behavior of the associated target. For interaction recognition, we first collect a set of 335 interactions (\ie, verbs), among them 327 from WordNet\cite{fellbaum1998wordnet}. Note that, in our interaction set, the same verb may have different meanings and thus is assigned to different interaction types, \eg, {\tt hold(v.01)} for ``being the physical support of someone'' and {\tt hold(v.02)} for ``someone in one’s hands or arm'', for precise relationship description. Due to limited space, we refer reader to \textbf{supplementary material} for the full set of interaction types. After this, we provide an directed interaction label, represented with a triplet of <\emph{subject}, \emph{predicate}, \emph{object}> (see Fig.~\ref{fig:smot} (b)), between two trajectories if associated objects are interacting. Finally, for the overall video captioning, we utilize a concise text summary to describe object trajectories in the video from a global perspective.

To ensure the high-quality annotations, we adopt a multi-round mechanism. Specifically, each video will be first manually annotated by a few volunteers who are familiar with the task and an expert working on related problems. After this initial round, all annotations of the video will be sent to a validation team former by more than two experts for inspection. If initial annotations are not unanimously agreed by the experts, they will be returned to the original labeling team for refinement using the feedback from the validation team. We repeat this procedure for each video until the annotations of all video sequences in BenSMOT are completed. An annotation example in our BenSMOT is shown in Fig.~\ref{fig:smot} (b), and more can be found in \textbf{supplementary material}.

\begin{figure*}[t!]
    \centering
    \begin{minipage}[b]{0.32\linewidth}
        \includegraphics[width=1\linewidth]{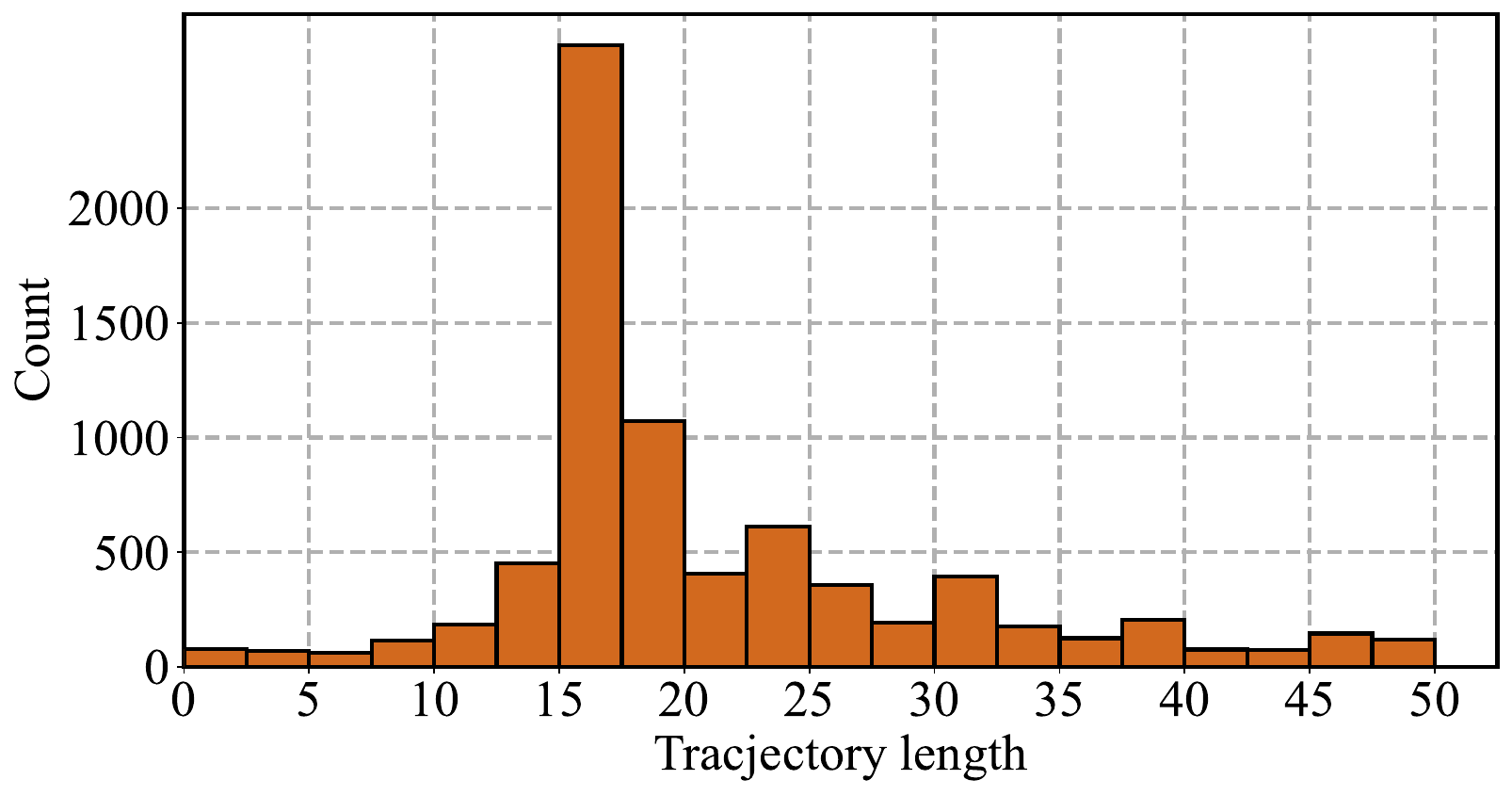}
        \captionsetup{labelformat=empty, font={scriptsize}, skip=2pt}
        \subcaption*{(a) Trajectory length distribution}
        \label{subfig:tracj length}
    \end{minipage}
    \begin{minipage}[b]{0.32\linewidth}
        \includegraphics[width=1\linewidth]{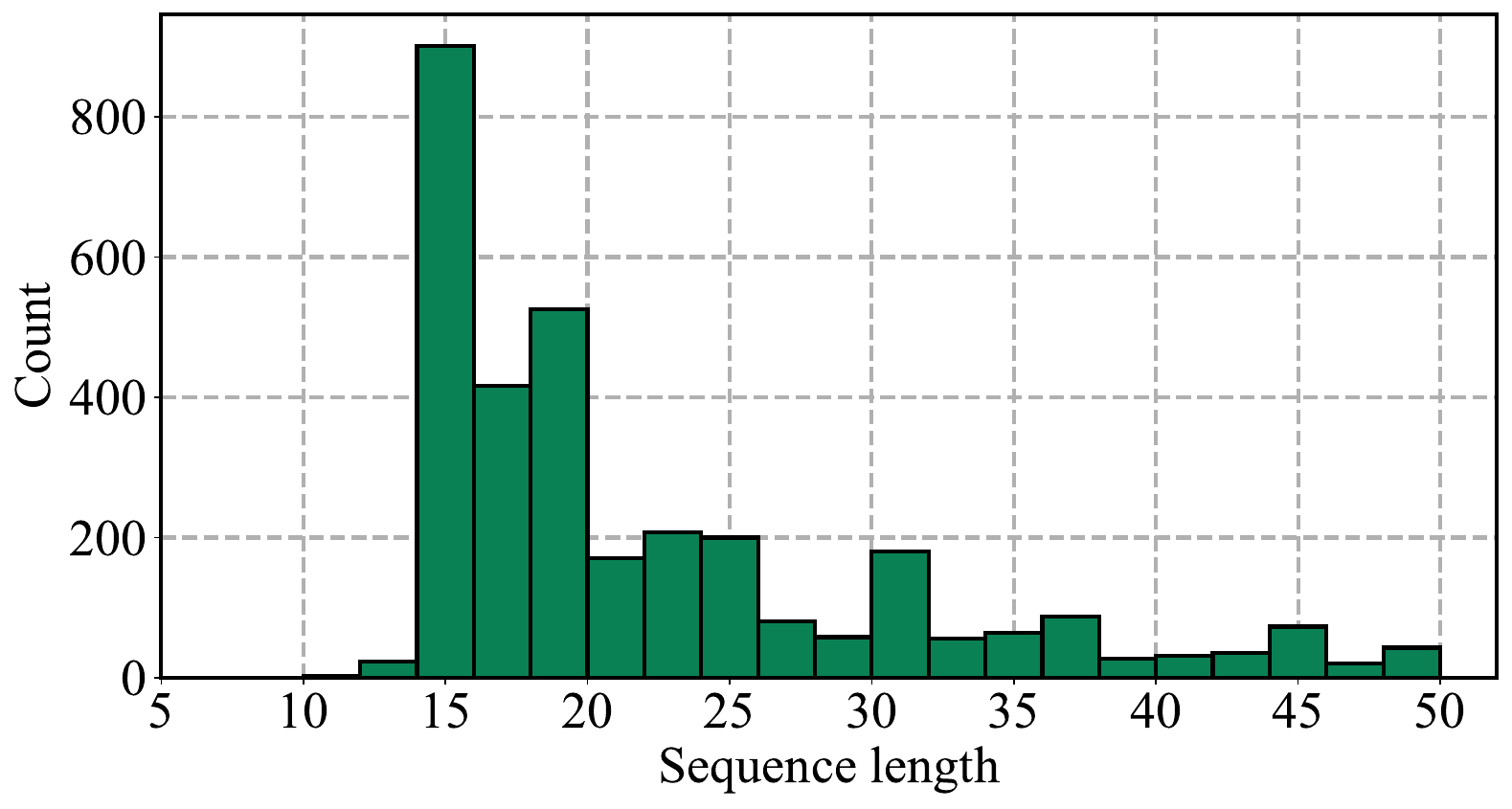}
        \captionsetup{labelformat=empty, font={scriptsize}, skip=2pt}
        \subcaption*{(b) Sequence length distribution}
        \label{subfig:seq length}
    \end{minipage}
    \begin{minipage}[b]{0.32\linewidth}
        \includegraphics[width=1\linewidth]{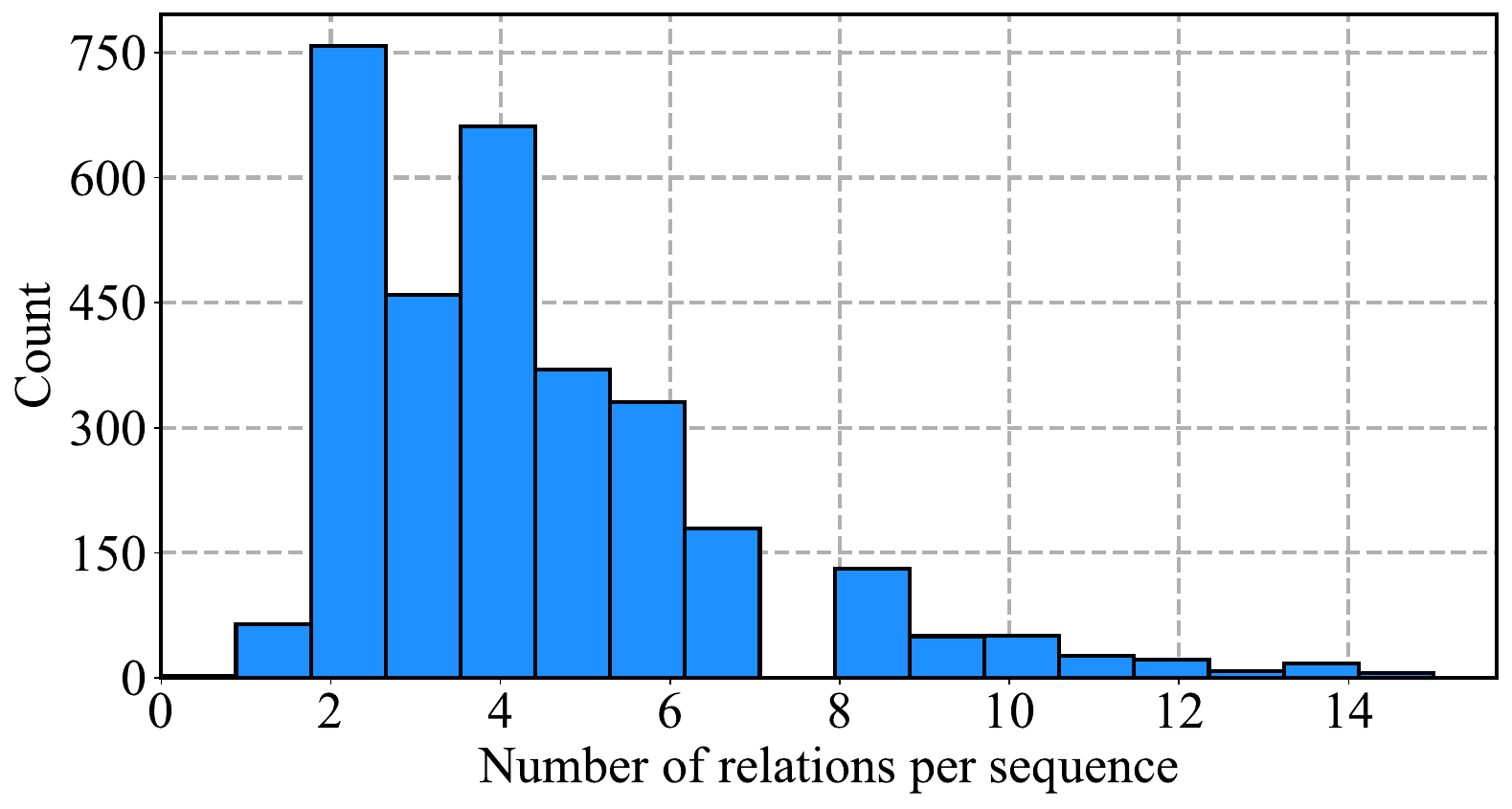}
        \captionsetup{labelformat=empty, font={scriptsize}, skip=2pt}
        \subcaption*{(c) Interactions count distribution}
        \label{subfig:relation num}
    \end{minipage}
    
    \begin{minipage}[b]{0.32\linewidth}
        \includegraphics[width=1\linewidth]{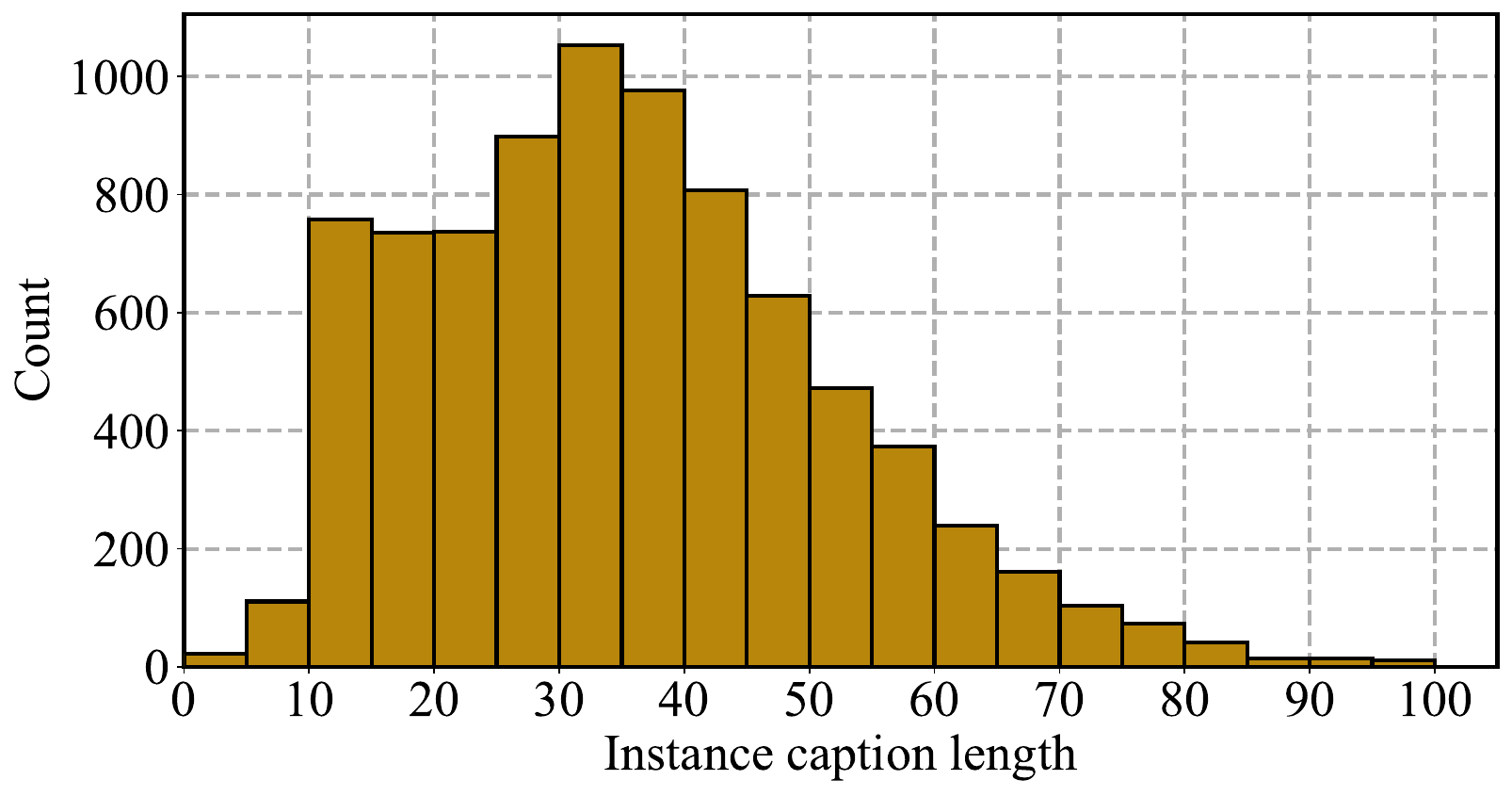}
        \captionsetup{labelformat=empty, font={tiny}, skip=2pt}
        \subcaption*{(d) Inst caption length distribution}
        \label{subfig:instance caption}
    \end{minipage}
    \begin{minipage}[b]{0.32\linewidth}
        \includegraphics[width=0.98\linewidth]{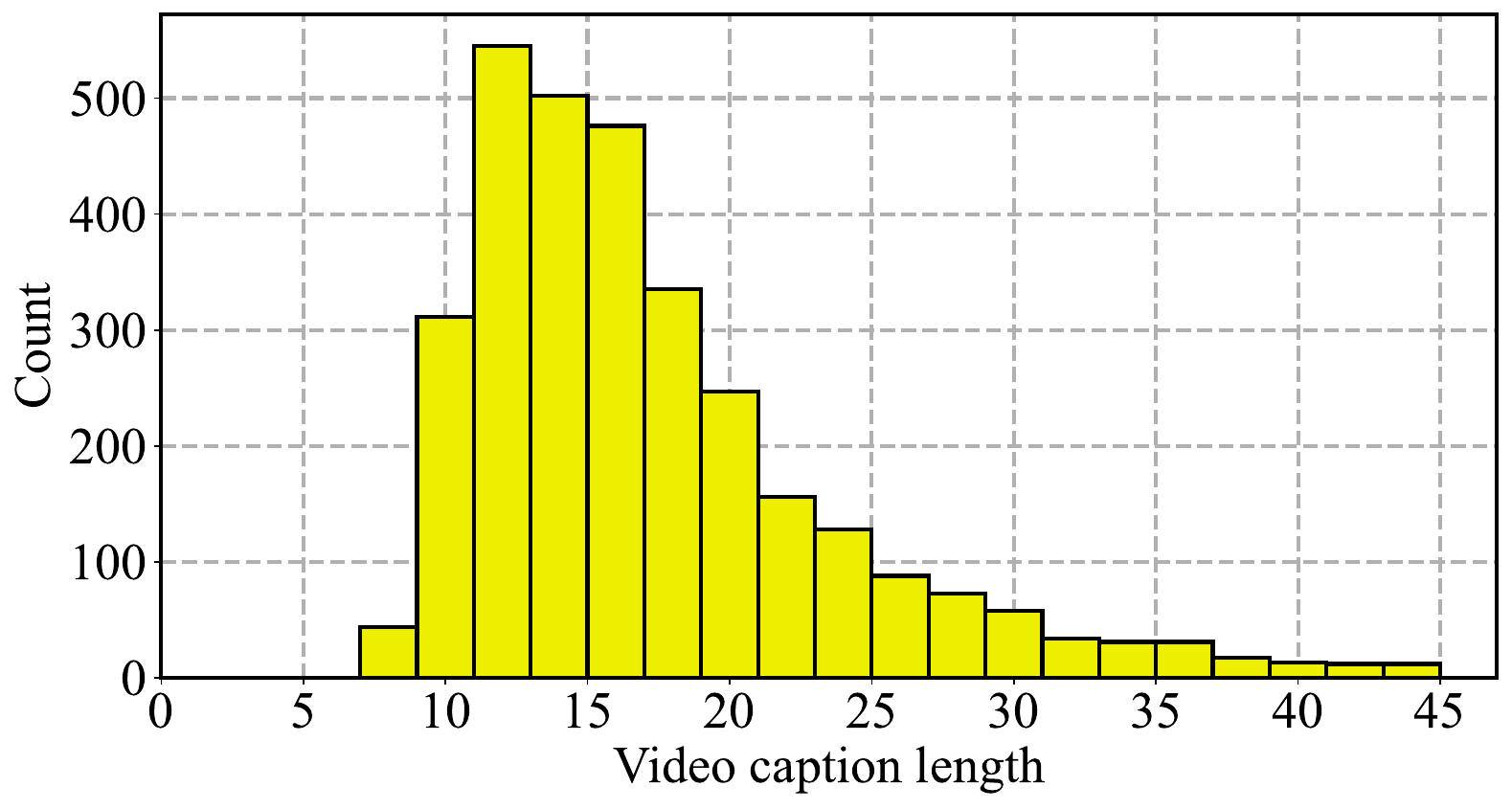}
        \captionsetup{labelformat=empty, font={tiny}, skip=2pt}
        \subcaption*{(e) Video caption length distribution}
        \label{subfig:video caption}
    \end{minipage}
    \begin{minipage}[b]{0.32\linewidth}
        \includegraphics[width=1\linewidth, height=58pt]{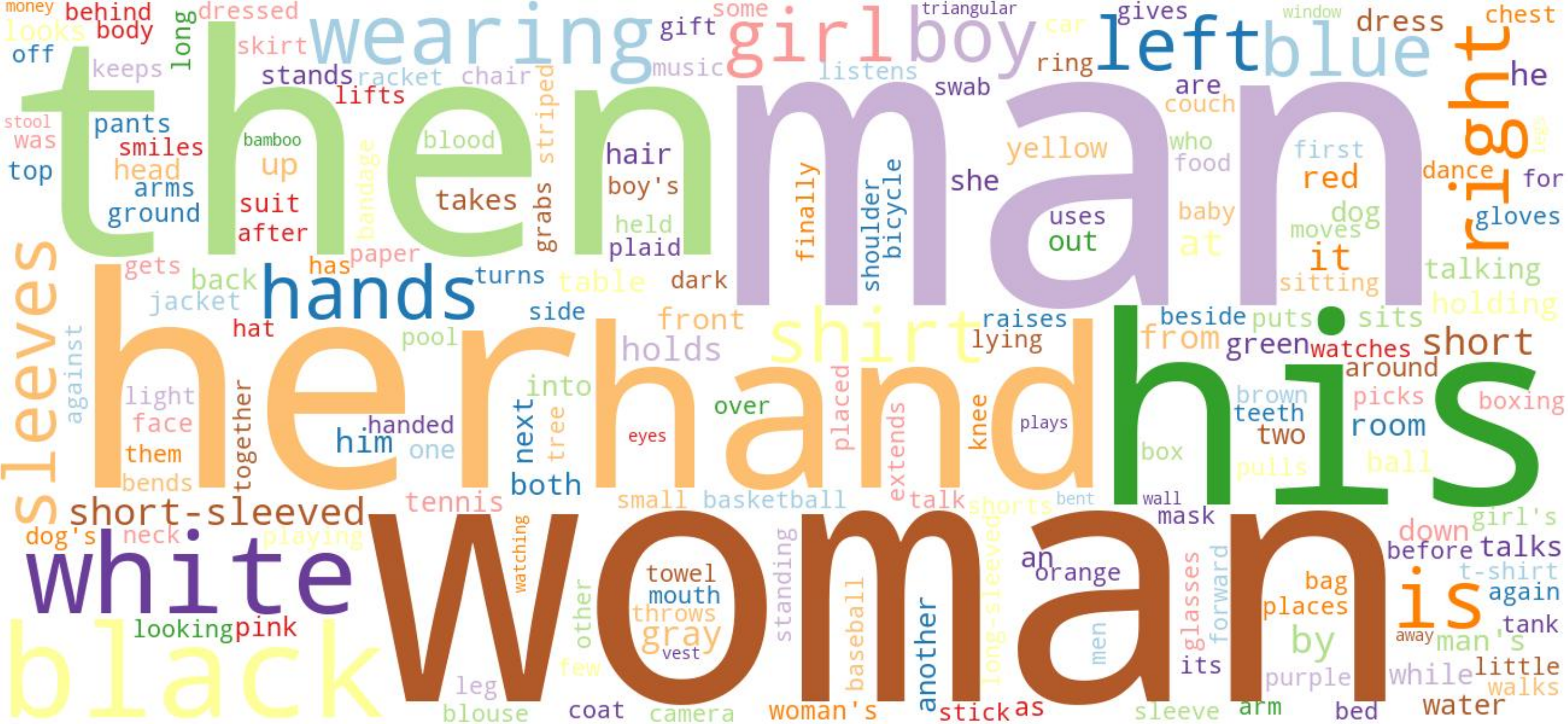}
        \captionsetup{labelformat=empty, font={scriptsize}, skip=2pt}
        \subcaption*{(f) WordCloud of caption words}
        \label{subfig:wordcloud}
    \end{minipage}

    \caption{Representative statistics on BenSMOT, including distributions of target trajectory length (in seconds) in (a), sequence length (in seconds) in (b), number of interactions per sequence in (c), instance caption length in (d), video caption length in (e), and wordcloud of all caption words with prepositions excluded in (f).}
    \label{fig:statistic}
\end{figure*}

\noindent
\textbf{Statistics of annotation.} In order to better understand BenSMOT, we show some representative statics of annotations in Fig.~\ref{fig:statistic}. Specifically, we demonstrate the distributions of object trajectory length, video length, number of instance interactions, instance caption length, video caption length, and wordcloud. From Fig.~\ref{fig:statistic} (c), it is worth noting that different targets in videos have frequent interactions. In addition, for the instance captions in Fig.~\ref{fig:statistic} (d), we provide relatively longer textual descriptions, which allows precise understanding.

\subsection{Dataset Split and Evaluation Metric}
\label{subsec:label and metric}

\textbf{Training/Test Split.} BenSMOT contains 3,292 videos, captured from 47 different scenarios. Within each scenario, we use 70\% of the sequences for training, and the rest 30\% for test. During dataset split, we try to keep the distributions of training and test sets similar. Eventfully, the training set of BenSMOT comprises 2,284 sequences with 104K frames, and the test set consists of 1,008 videos with 47K frames. Please see \textbf{supplementary material} for more split details.

\noindent
\textbf{Evaluation Metric.} We apply multiple metrics on BenSMOT for evaluation. Specifically, to assess the performance in object trajectory estimation, we employ higher order tracking accuracy (HOTA), association accuracy (AssA), detection accuracy (DetA), and localization accuracy (LocA) by following~\cite{luiten2021hota}, CLEAR metrics~\cite{bernardin2008evaluating} including multiple object tracking accuracy (MOTA), false positives (FP), false negatives (FN), and ID switches (IDs), and ID metrics~\cite{ristani2016performance} containing identification precision (IDP), identification recall (IDR) and related F1 score (IDF1). For instance and video captioning, we follow existing methods (\eg,~\cite{lin2022swinbert,shen2023accurate}) and employ BLEU \cite{papineni2002bleu}, ROUGE \cite{lin2004rouge}, METEOR \cite{banerjee2005meteor}, and CIDEr \cite{vedantam2015cider} for thorough evaluation. Lastly, to assess performance of interaction recognition, we leverage classic metrics such as Precision (Prcn), Recall (Rcll), and F1, which are widely used in the context of classification tasks. 


\begin{figure}[!t]
  \centering
  \includegraphics[width=0.9\linewidth]{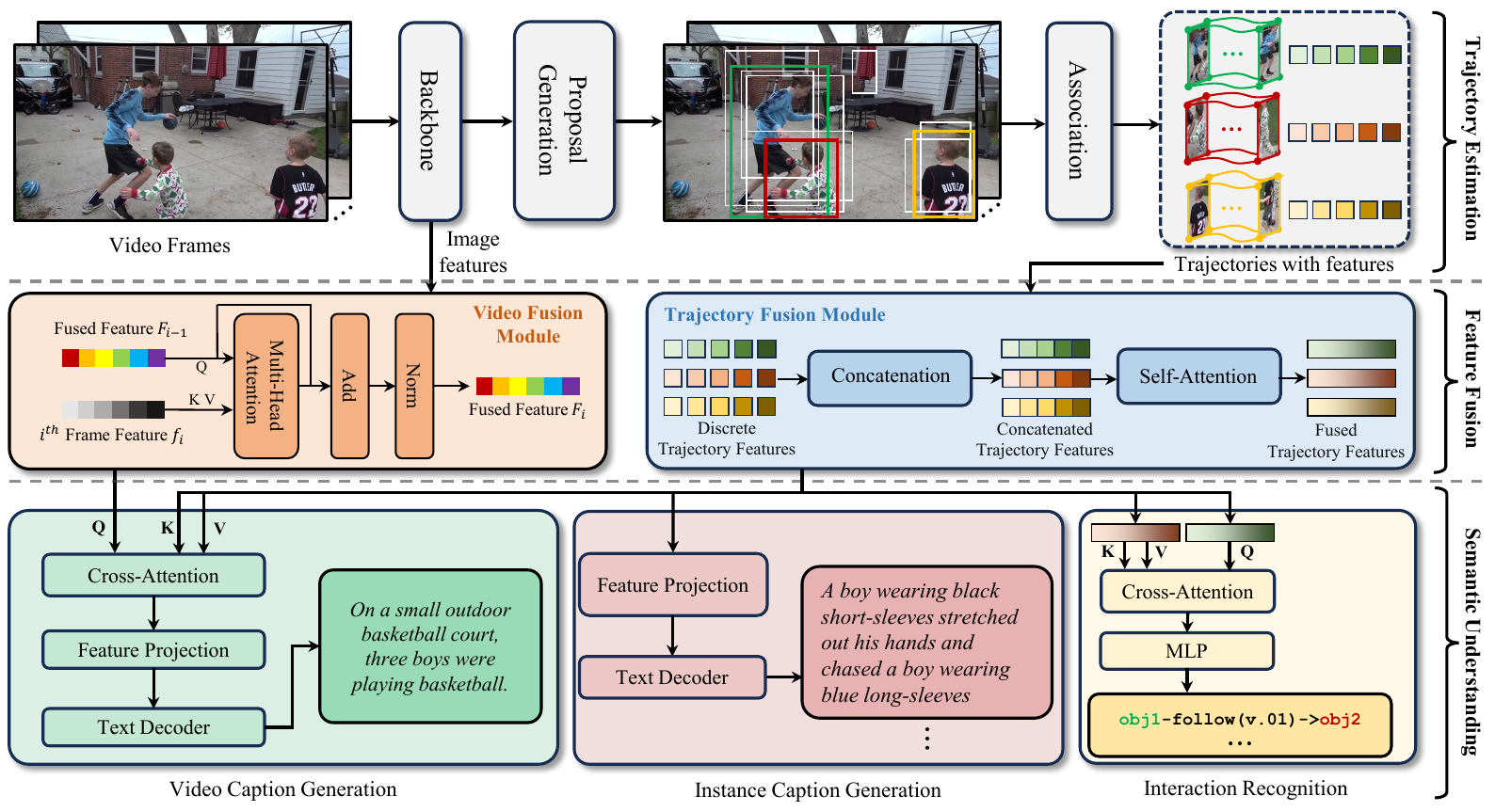}
  \caption{Illustration of SMOTer, which contains three components of trajectory estimation for tracking, feature fusion, and trajectory-associated semantic understanding.}
  \label{fig:overview}
\end{figure}

\section{Methodology: A New Baseline for SMOT}

\textbf{Overview.} To encourage development of SMOT algorithms, we propose a simple yet effective tracker on BenSMOT, dubbed SMOTer. As illustrated in Fig.~\ref{fig:overview}, SMOTer logically consists of three key components. The first component, as detailed in Sec~\ref{subsec:object tracking}, is dedicated to object trajectory estimation for tracking, which is the basis for SMOT. Then, the second component, as explained in Sec~\ref{subsec:feature fusion}, focuses on aggregating features from each frame into the overall video feature and meanwhile merging features of individual objects into target trajectory features using two distinct fusion modules, for subsequent semantic understanding. Finally, the third component, as described in Sec~\ref{subsec:semantic}, aims to predict trajectory-associated semantic details in the video, including instance captions, instance interactions, and overall video caption. Please notice that, our SMOTer is specially designed for SMOT and trained in an end-to-end manner, as in Sec~\ref{subsec:end-to-end}.

\subsection{Target Trajectory Estimation for Tracking}
\label{subsec:object tracking}
Give a video with $N$ frames, SMOTer first uses a CNN backbone~\cite{yu2018deep} to extract their features $\{f_i\}_{i=1}^{N}$, where $f_i$ denotes the feature of the $i^{\text{th}}$ frame. Then, an initial set of proposals is generated using proposal generator of a popular detection architecture~\cite{duan2019centernet} with threshold filtering by $\tau_p$. Afterwards, all reserved proposals are send to the association module (BYTE~\cite{zhang2022bytetrack} is used in SMOTer for association) to obtain object trajectories. Finally, leveraging target coordinates within the trajectories and image features, target features are acquired for each target in trajectories through RoI Pooling~\cite{girshick2015fast}. Specifically, given feature $f_i$ of the $i^{th}$ frame and target coordinates $B_j^i$ corresponding to the $j^{th}$ object trajectory in this frame, we extract the corresponding target feature $t_j^i = {\tt RoIPooling}(f_i, B_j^i)$ and apply it in subsequent feature fusion for semantic understanding. 

In summary, at this stage, SMOTer predicts object trajectories for tracking in a given video and extracts image and target features for subsequent stages.

\subsection{Feature Fusion}
\label{subsec:feature fusion}
Given image and target features from previous stage in Sec.~\ref{subsec:object tracking}, we design two feature fusion modules, including attention-based Video Fusion Module (VFM) and concatenation-based Trajectory Fusion Module (TFM). VFM aims to merge image-level features into overall video features while TFM focuses on integrating target features from each frame into target trajectory features.

\noindent
\textbf{Video Fusion Module.} VFM takes features $\{f_i\}_{i=1}^{N}$ of $N$ frames as input and sequentially feeds them into a cross-attention module for video feature fusion. In specific, the fused video feature $F_i$ up to the $i^{\text{th}}$ ($i>1$) frame is computed as 
\begin{equation} \label{eq1}
    F_i = {\tt CA}(F_{i-1}, f_i) \;\;\;\;\; i>1
\end{equation}
where ${\tt CA}(\textbf{z},\textbf{u})$ denotes cross-attention with $\textbf{z}$ generating query and $\textbf{u}$ key/value as in~\cite{vaswani2017attention}. For the first frame ($i=1$), $F_1=f_1$. Finally, with the help of VFM, we can generate the overall fused video feature $\Tilde{F} = F_N$.

\noindent
\textbf{Trajectory Fusion Module.} Unlike VFM, TFM is to integrate multiple target features belonging to the same object trajectory, which are distributed across different frames, into a single trajectory feature. Concretely, given $t_i^j$ for object feature of trajectory $j$ in frame $i$, we perform average pooing to yield a relatively simple 2D feature $\hat{t}_i^j={\tt AvgPooling}(t_{i}^{j})$, where $\hat{t}_i^j \in \mathbb{R}^{4 \times d}$ with dimension $d=256$. Then, for the $j^{\text{th}}$ object trajectory, we first concatenate its discrete features into an integrated one $\hat{T_j} = {\tt Concat}(\hat{t}_1^j, \hat{t}_2^j,\cdots, \hat{t}_{N_{j}}^{j})$, where $N_j$ denotes trajectory length. After that, self-attention is used to enhance trajectory feature as follows
\begin{equation}
    \Tilde{T_j} = {\tt SA}(\hat{T_j})
\end{equation}
where ${\tt SA}(\textbf{z})$ denotes self-attention with $\textbf{z}$ generating query/key/value as in~\cite{vaswani2017attention}, and $\Tilde{T_j}$ is the enhanced feature for trajectory $j$.

\subsection{Semantic Understanding}
\label{subsec:semantic}
In the third phase, SMOTer applies the fused video feature $\Tilde{F}$ and trajectory features $\Tilde{T} = \{\Tilde{T_1}, \Tilde{T_2}, ..., \Tilde{T_M}\}$, where $M$ denotes the number of trajectories in the video, to tackle  trajectory-associated semantic understanding tasks in SMOT, \ie, video captioning, instance captioning, and interaction recognition. 

\noindent
\textbf{Video captioning.} Unlike current video captioning, in SMOTer we incorporate trajectory features into describing video content and allow more comprehensive human-centric caption generation. Specifically, SMOTer sequentially injects trajectory features $\Tilde{T}$ into video feature $\Tilde{F}$ with cross-attention and then predicts captioning result $R_{\text{Vid}}$ via a linear project layer and a text decoder, as follows
\begin{equation} \label{eq2}
   R_{\text{Vid}} = {\tt Dec}_{\text{Vid}}({\tt Linear}({\tt CA}(\Tilde{F}, \Tilde{T})))
\end{equation}
where ${\tt Dec}_{\text{Vid}}(\cdot)$ is a video text decoder for video captioning (please kindly refer to \textbf{supplementary material} for details), and ${\tt Linear}(\cdot)$ is a linear projection.

\noindent
\textbf{Instance captioning.} Different from video captioning, for instance captioning, we directly generate the results based on the target trajectory features. Specifically, given the feature $\Tilde{T_j}$ for the $j^{\text{th}}$ ($1\leq j\leq M$) trajectory, we apply a linear projection layer followed by the designed text decoder to predict the captioning result $R_{\text{Ins}}^{j}$ of each trajectory, depicted as follows,
\begin{equation} \label{eg3}
    R_{\text{Ins}}^{j} = {\tt Dec}_{\text{Ins}}({\tt Linear}(\Tilde{T_j})) \;\;\;\;\; (1\leq j\leq M)
\end{equation}
where ${\tt Dec}_{\text{Ins}}(\cdot)$ is instance text decoder~\cite{yang2023vid2seq} for instance captioning. Please refer to \textbf{supplementary material} for more details because of space limitation.

\noindent
\textbf{Interaction recognition.} Besides captioning, SMOTer predicts the interaction between trajectories, which is crucial for video understanding. Particularly, in SMOTer we concentrate solely on interactions between two trajectories, as our current focus doesn't extend to understanding interactions involving three or more targets, leaving it for future research. More specifically, for two trajectories $j$ and $k$, we assume the $j^{th}$ trajectory in a video sequence is the active trajectory and the $k^{th}$ trajectory the passive trajectory. Then, we first fuse the features of active and passive trajectories using cross-attention and then predict the interaction result $R_{\text{Int}}^{j,k}$ using multi-layer perception~\cite{taud2018multilayer}, as follows,
\begin{equation} \label{eq4}
    R_{\text{Int}}^{j,k} = {\tt MLP}({\tt CA}(\Tilde{T_j}, \Tilde{T_k})),
\end{equation}
where ${\tt MLP}(\cdot)$ denotes the multi-layer perception. Please note, interchanging the active and passive trajectories may yield disparate outcomes, \ie, $R_{\text{Int}}^{j,k}$ and $R_{\text{Int}}^{k,j}$ may exhibit dissimilar characteristics, as indicated by the formula above.

\subsection{End-to-end Training}
\label{subsec:end-to-end}
End-to-end training of SMOTer is not easy due to conflict between two types of training requirements, \ie, the trajectory estimation is usually trained frame-by-frame, while the captioning tasks, and interaction recognition are often trained at the entire video level, resulting in two different losses in training. In addition, due to the frame-by-frame training nature of trajectory estimation for tracking, the complete sequence is hard to be input at once during training, preventing the model from obtaining full trajectories thus hindering subsequent caption generation and interaction prediction. To address these issues, we draw inspiration from~\cite{zhou2022global} and adopt a strategy of completing full trajectory association after detection on the entire video. Specifically, we first compute detection loss after each frame input and then perform trajectory association after the entire sequence is detected. Subsequently, with complete trajectories, we compute losses for trajectory-associated instance captioning, interaction recognition, and video captioning tasks, realizing end-to-end training of SMOTer. For the detailed training loss of SMOTer, please kindly refer to the \textbf{supplementary material}.

\section{Experiments}

\textbf{Experimental setup.} We conduct experiments with 4 Nvidia Tesla V100 GPUs (32GB). The batch size is 1 per GPU. Each batch has a video clip with multiple frames. We use the AdamW optimizer with an initial learning rate of $5.0\times10^{-4}$. During training, we filter out proposals with scores lower than the threshold $\tau_{p} = 0.3$. For lost tracklets, we retain them for 30 frames in case they reappear. Due to memory constraints and limited accuracy of trajectory estimation in early training stages, we sample 6-8 frames per sequence for training detection model and only compute tracking related losses in the first 20K training iterations.

As SMOT is new and no suitable methods exist for comparison with SMOTer, we logically divide SMOT into two components: trajectory estimation for tracking and semantic understanding (including instance/video captioning, and interaction recognition). Subsequently, we devise a series of two-stage models based on existing MOT frameworks. Leveraging several state-of-the-art and classic MOT models such as SORT~\cite{bewley2016simple}, DeepSORT~\cite{wojke2017simple}, OC-SORT~\cite{cao2023observation}, ByteTrack~\cite{zhang2022bytetrack}, TransTrack~\cite{sun2020transtrack}, MOTR~\cite{zeng2022motr}, and MOTRv2~\cite{zhang2023motrv2} as foundations, we integrate feature fusion and semantic generation functionalities akin to SMOTer and conduct comparative experiments on BenSMOT. To ensure fair experiments, we employ the same training framework for comparative experiments of two-stage approaches. Additionally, for models based on two-stage MOT methods like SORT, we use the same detector~\cite{duan2019centernet} as SMOTer during training and evaluation.

\subsection{Comparison on Trajectory Estimation for Tracking.}

We first compare SMOTer with other MOT methods for trajectory estimation on BenSMOT. Specifically, SMOTer is fully end-to-end trained, while other two-stage MOT models are trained with their own strategies, concentrating solely on tracking. As depicted in Tab.~\ref{tab:comparison-tracking}, despite being at a certain degree of natural disadvantage, our SMOTer achieves comparable or even superior performance in tracking compared to state-of-the-art MOT models. SMOTer consistently achieves top-tier performance across nearly all metrics, demonstrating impressive results in key metrics. For instance, it achieves a notable 71.98$\%$ in HOTA, 77.71$\%$ in MOTA, and 80.65$\%$ in IDF1. This suggests that subsequent training tasks do not degrade tracking performance, validating effectiveness of our end-to-end training strategy. In fact, by comparing with the standard ByteTrack, we observe that although SMOTer adopts a highly similar structure to ByteTrack in tracking, it achieves superior performance, showing +3.15$\%$ in HOTA and +3.74$\%$ in MOTA. Moreover, even in metrics where it does not excel, such as FP and LocA, SMOTer exhibits significant improvements compared to ByteTrack, implying that semantic understanding tasks can reciprocally aid in tracking. 

\begin{table*}[!t]\scriptsize
    \centering
    \renewcommand{\arraystretch}{1}
    \tabcolsep=1.2mm
    \caption{Comparison between SMOTer and two-stage MOT methods regarding tracking performance on BenSMOT. The best two results are shown in \textcolor{red}{red} and \textcolor{blue}{blue} fonts.}
    \resizebox{0.9\textwidth}{!}{
    \begin{tabular}{r|ccccccccccc}
        \toprule[1.2pt]
        \textbf{Method} & \textbf{HOTA}$\uparrow$ & \textbf{AssA}$\uparrow$ & \textbf{DetA}$\uparrow$ & \textbf{LocA}$\uparrow$ & \textbf{MOTA}$\uparrow$ & \textbf{FN}$\downarrow$ & \textbf{FP}$\downarrow$ & \textbf{IDs}$\downarrow$ & \textbf{IDR}$\uparrow$ & \textbf{IDP}$\uparrow$ & \textbf{IDF1}$\uparrow$\\
        \midrule[0.8pt]
        SORT~\cite{bewley2016simple} & 48.49 & 38.95 & 60.91 & 87.50 & 53.58 & 24001 & \textcolor{blue}{5105} & 13875 & 60.85 & 48.43 & 53.93 \\
        DeepSORT~\cite{wojke2017simple} & 50.12 & 40.23 & 61.45 & \textcolor{blue}{87.67} & 54.29 & 22890 & 5540 & 11278 & 62.10 & 51.11 & 56.76 \\
        OC-SORT~\cite{cao2023observation} & 51.00 & 41.42 & 63.31 & 87.61 & 55.19 & 21061 & 5388  & 15049 & 63.92 & 53.10 & 58.01 \\
        ByteTrack~\cite{zhang2022bytetrack} & 68.84 & 71.15 & 67.10 & 85.15 & 73.87 & \textcolor{blue}{15419} & 7070 & 1712 & 82.25 & \textcolor{blue}{74.83} & 78.37 \\
        TransTrack~\cite{sun2020transtrack} & \textcolor{blue}{71.31} & 73.34 & \textcolor{blue}{69.67} & \textcolor{red}{91.31} & \textcolor{blue}{74.08} & 20124 & \textcolor{red}{4420} & 2530 & \textcolor{red}{85.63} & 72.75 & \textcolor{blue}{78.67} \\
        MOTR~\cite{zeng2022motr} & 66.10 & 73.12 & 55.14 & 86.30 & 45.19 & 31297 & 11178 & \textcolor{blue}{617} & 72.39 & 70.12 & 68.97 \\
        MOTRv2~\cite{zhang2023motrv2} & 65.28 & \textcolor{red}{76.82} & 51.30 & 86.09 & 45.52 & 40765 & 20923 & \textcolor{red}{430} & 78.47 & 65.51 & 70.76 \\
        \midrule[0.8pt]
        SMOTer (ours) & \textcolor{red}{71.98} & \textcolor{blue}{73.71} & \textcolor{red}{70.79} & 87.11 & \textcolor{red}{77.71} & \textcolor{red}{12534} & 6388 & 1702 & \textcolor{blue}{83.82} & \textcolor{red}{77.97} & \textcolor{red}{80.65} \\
        \bottomrule[1.2pt]
    \end{tabular}
    \label{tab:comparison-tracking}
    }
\end{table*}


\begin{table*}[!t]
    \centering
    \renewcommand{\arraystretch}{1}
    \tabcolsep=1.0mm
    \caption{Comparison of SMOTer against two-stage methods based on MOT models regarding semantic understanding. The best two results are highlighted in \textcolor{red}{red} and \textcolor{blue}{blue}.}
    \resizebox{0.9\textwidth}{!}{
    \begin{tabular}{r|cccc|cccc|ccc}
        \toprule[1.2pt]
            & \multicolumn{4}{|c|}{\textbf{Video Caption}} & \multicolumn{4}{|c|}{\textbf{Instance Caption}} & \multicolumn{3}{|c}{\textbf{Interaction}} \\
            \midrule[0.8pt]
        \textbf{Method} & \textbf{BLEU}$\uparrow$ & \textbf{ROUGE}$\uparrow$ & \textbf{METEOR}$\uparrow$ & \textbf{CIDEr}$\uparrow$ & \textbf{BLEU}$\uparrow$ & \textbf{ROUGE}$\uparrow$ & \textbf{METEOR}$\uparrow$ & \textbf{CIDEr}$\uparrow$ & \textbf{Prec}$\uparrow$ & \textbf{Rcll}$\uparrow$ & \textbf{F1}$\uparrow$\\
        \midrule[0.8pt]
            SORT~\cite{bewley2016simple} & \textcolor{blue}{0.245} & 0.224 & 0.202 & 0.298 & 0.233 & \textcolor{red}{0.245} & 0.208 & 0.056 & 0.363 & 0.259 & 0.302 \\
            DeepSORT~\cite{wojke2017simple} & 0.198 & 0.213 & 0.187 & \textcolor{blue}{0.309} & 0.238 & 0.212 & 0.199 & 0.065 & 0.365 & 0.277 & 0.310 \\
            OC-SORT~\cite{cao2023observation} & 0.231 & 0.252 & 0.215 & 0.242 & 0.270 & 0.205 & 0.180 & 0.033 & 0.384 & 0.291 & 0.331 \\
            ByteTrack~\cite{zhang2022bytetrack} & 0.224 & 0.225 & 0.212 & 0.266 & \textcolor{blue}{0.304} & \textcolor{blue}{0.242} & \textcolor{red}{0.224} & 0.064 & \textcolor{red}{0.443} & 0.258 & 0.326 \\
            TransTrack~\cite{sun2020transtrack} & \textcolor{red}{0.247} & 0.248 & 0.209 & 0.269 & 0.283 & 0.219 & 0.201 & \textcolor{blue}{0.074} & 0.406 & 0.311 & \textcolor{red}{0.376} \\
            MOTR~\cite{zeng2022motr} & 0.187 & 0.254 & 0.203 & 0.244 & 0.230 & 0.209 & 0.182 & 0.061 & 0.425 & 0.314 & 0.354 \\
            MOTRv2~\cite{zhang2023motrv2} & 0.217 & \textcolor{blue}{0.258} & \textcolor{blue}{0.219} & 0.248 & 0.238 & 0.241 & 0.204 & 0.059 & 0.313 & \textcolor{red}{0.395} & 0.349 \\
            \midrule[0.8pt]
            SMOTer (ours) & \textcolor{blue}{0.245} & \textcolor{red}{0.261} & \textcolor{red}{0.223} & \textcolor{red}{0.343} & \textcolor{red}{0.306} & 0.223 & \textcolor{blue}{0.209} & \textcolor{red}{0.087} & \textcolor{blue}{0.434} & \textcolor{blue}{0.320} & \textcolor{blue}{0.368} \\
            \bottomrule[1.2pt]
    \end{tabular}
    \label{tab:comparison-generation}
    }
\end{table*}


\begin{table*}[!t]
    \centering
    \renewcommand{\arraystretch}{1}
    \tabcolsep=1.0mm
    \caption{Ablation experiments for evaluating various feature fusion strategies using results of three additional tasks. The best result is highlighted in \textcolor{red}{red}.}
    \resizebox{0.9\textwidth}{!}{
    \begin{tabular}{r|cccc|ccccccc}
        \toprule[1.2pt]
            & \multicolumn{4}{|c|}{\textbf{VFM (video fusion)}} & \multicolumn{7}{|c}{\textbf{TFM (trajectory fusion)}} \\
            \midrule[0.8pt]
        & \textbf{BLEU}$\uparrow$ & \textbf{ROUGE}$\uparrow$ & \textbf{METEOR}$\uparrow$ & \textbf{CIDEr}$\uparrow$ & \textbf{BLEU}$\uparrow$ & \textbf{ROUGE}$\uparrow$ & \textbf{METEOR}$\uparrow$ & \textbf{CIDEr}$\uparrow$ & \textbf{Prcn}$\uparrow$ & \textbf{Rcll}$\uparrow$ & \textbf{F1}$\uparrow$\\
        \midrule[0.8pt]
            Attention-based & 0.245 & \textcolor{red}{0.261} & \textcolor{red}{0.223} & 0.343 & 0.303 & 0.218 & 0.208 & 0.072 & 0.406 & \textcolor{red}{0.347} & 0.364 \\
            MLP-based & \textcolor{red}{0.251} & 0.223 & 0.200 & 0.220 & \textcolor{red}{0.337} & 0.219 & 0.205 & 0.072 & 0.380 & 0.252 & 0.303 \\
            Concatenation-based & 0.230 & 0.213 & 0.215 & \textcolor{red}{0.365} & 0.306 & \textcolor{red}{0.223} & \textcolor{red}{0.209} & \textcolor{red}{0.087} & \textcolor{red}{0.434} & 0.320 & \textcolor{red}{0.368} \\
            Addition-based & 0.140 & 0.157 & 0.104 & 0.127 & 0.182 & 0.137 & 0.114 & 0.041 & 0.228 & 0.163 & 0.205 \\
            \bottomrule[1.2pt]
    \end{tabular}
    \label{tab:fusion-ablation}
    }
\end{table*}

\subsection{Comparison on Semantic Understanding}
Besides tracking, we compare SMOTer with a set of two-stage models on their semantic understanding capabilities. Similarly, SMOTer undergoes complete end-to-end training. However, other two-stage models first train trackers on their own to acquire target trajectories and then use these trajectories to separately train the semantic understanding tasks (with the same feature fusion as SMOTer). From Tab.~\ref{tab:comparison-generation}, we can see SMOTer achieves highly favorable results on almost all metrics. Particularly, compared to the baseline method ByteTrack, SMOTer surpasses it on most metrics, including +2.1$\%$ BLEU and +7.7$\%$ CIDEr on video captioning, +2.3$\%$ CIDEr on instance captioning, and +4.2$\%$ F1 score on interaction recognition. More results can be seen in \textbf{supplementary material}.

\begin{figure*}[t!]
    \centering
    \begin{minipage}[b]{0.24\linewidth}
        \includegraphics[width=1\linewidth]{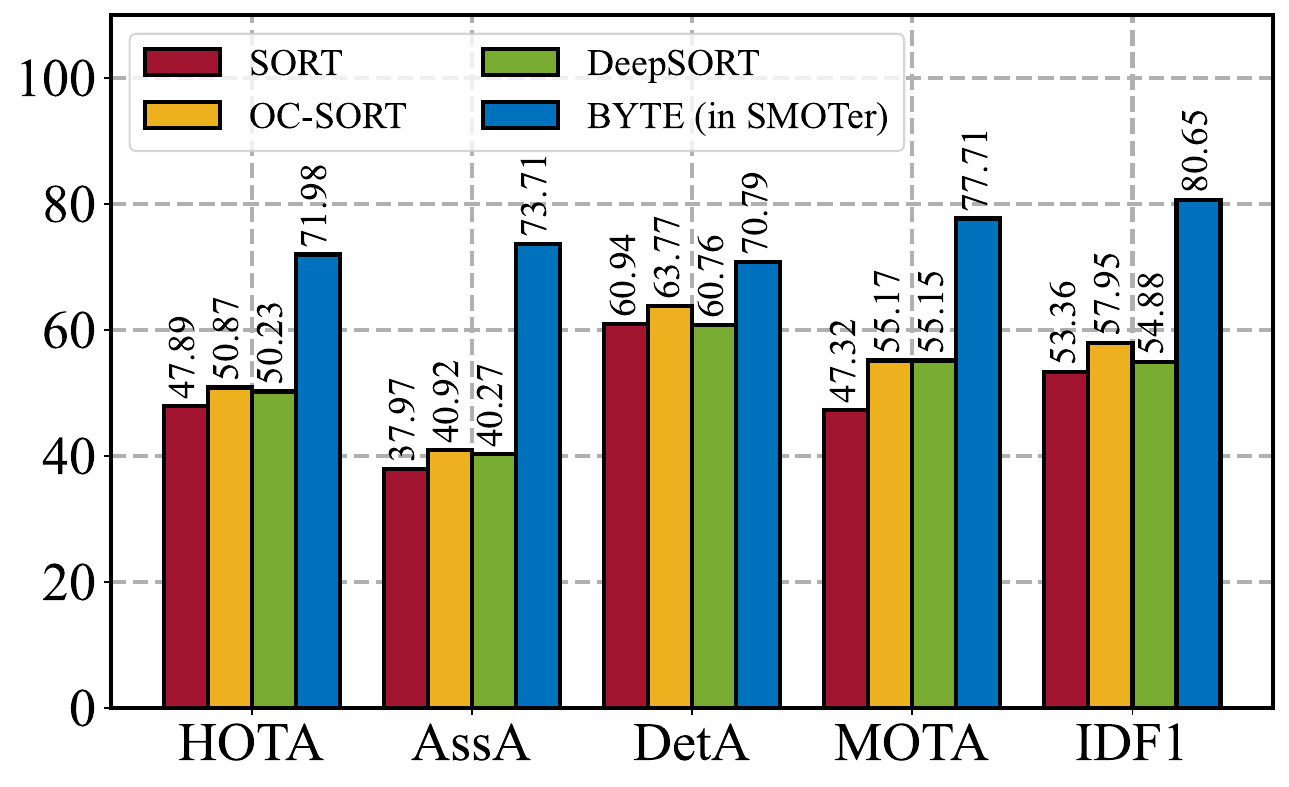}
        \captionsetup{labelformat=empty, font={scriptsize}, skip=2pt}
        \subcaption*{(a) Comparison results on object tracking}
    \end{minipage}
    \begin{minipage}[b]{0.24\linewidth}
        \includegraphics[width=1\linewidth]{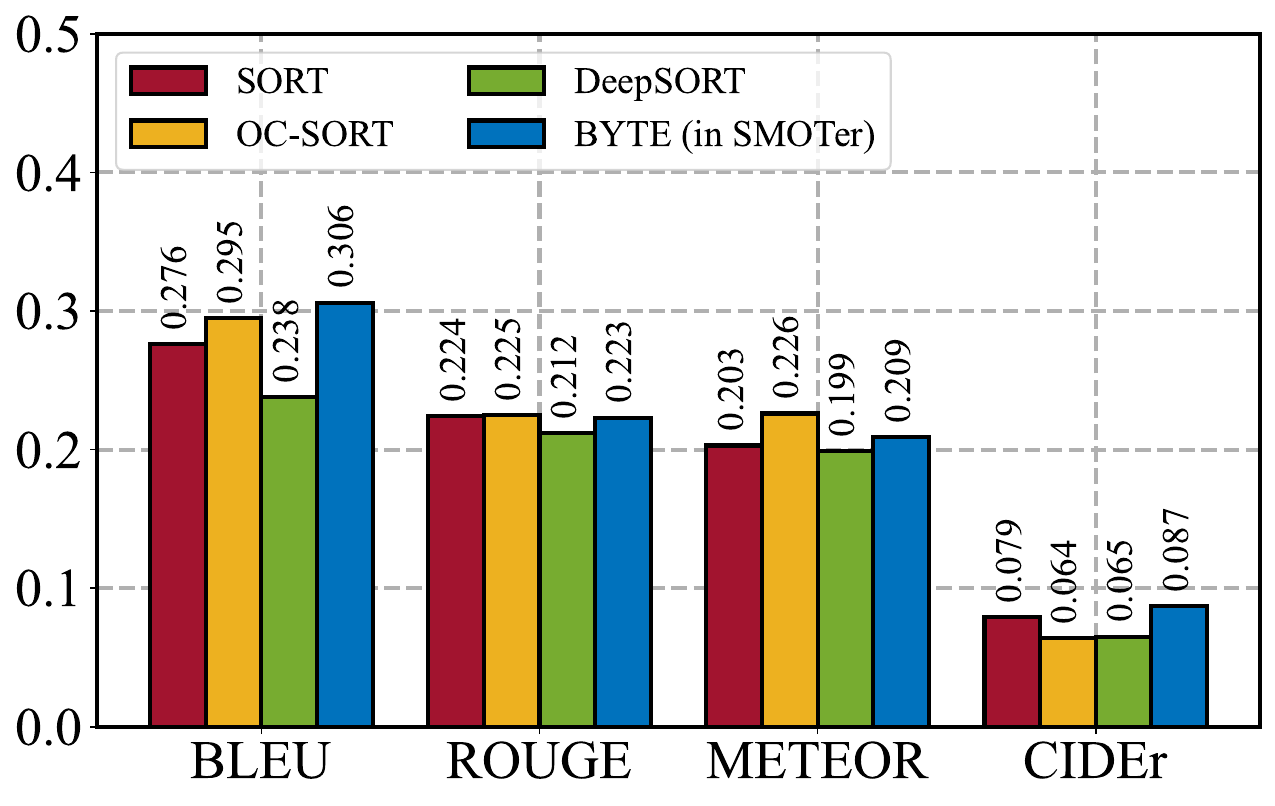}
        \captionsetup{labelformat=empty, font={scriptsize}, skip=2pt}
        \subcaption*{(b) Comparison results on video captioning}
    \end{minipage}
    \begin{minipage}[b]{0.24\linewidth}
        \includegraphics[width=1\linewidth]{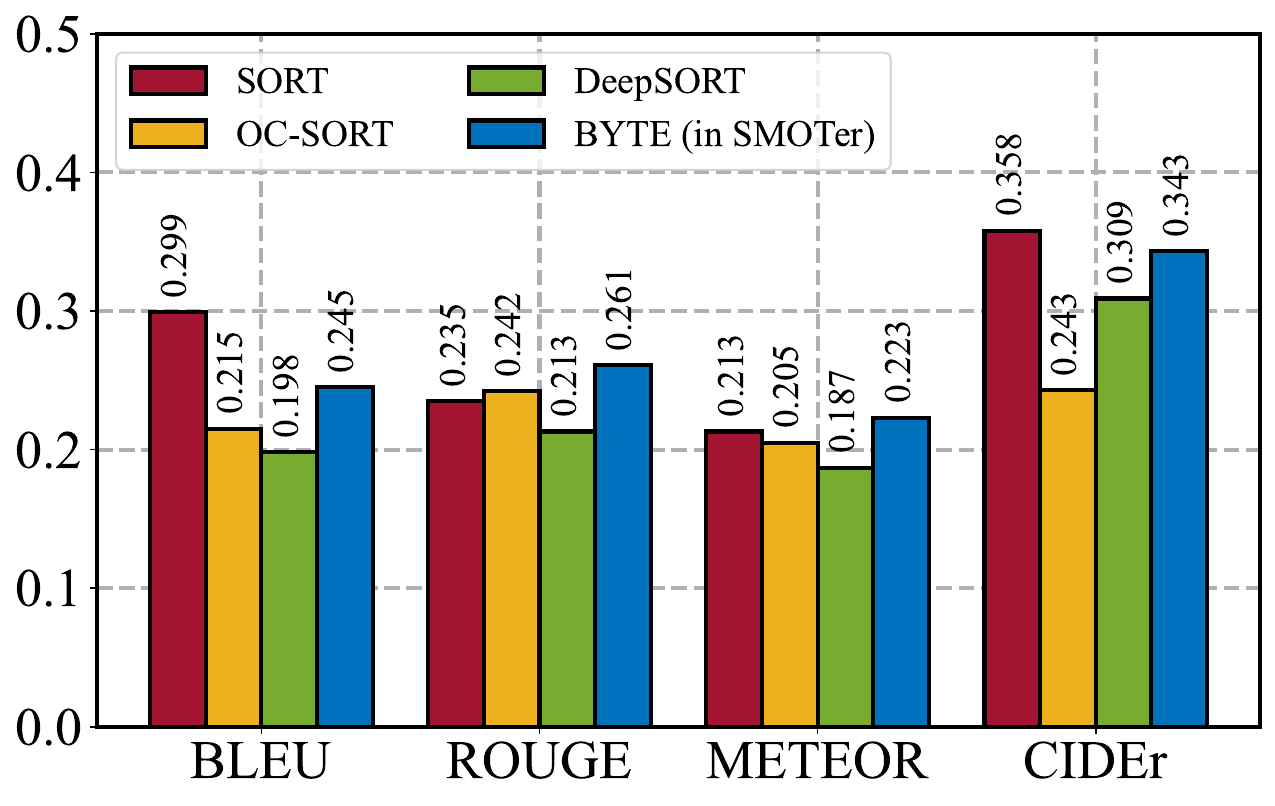}
        \captionsetup{labelformat=empty, font={scriptsize}, skip=2pt}
        \subcaption*{(c) Comparison results on instance captioning}
    \end{minipage}
    \begin{minipage}[b]{0.24\linewidth}
        \includegraphics[width=1\linewidth]{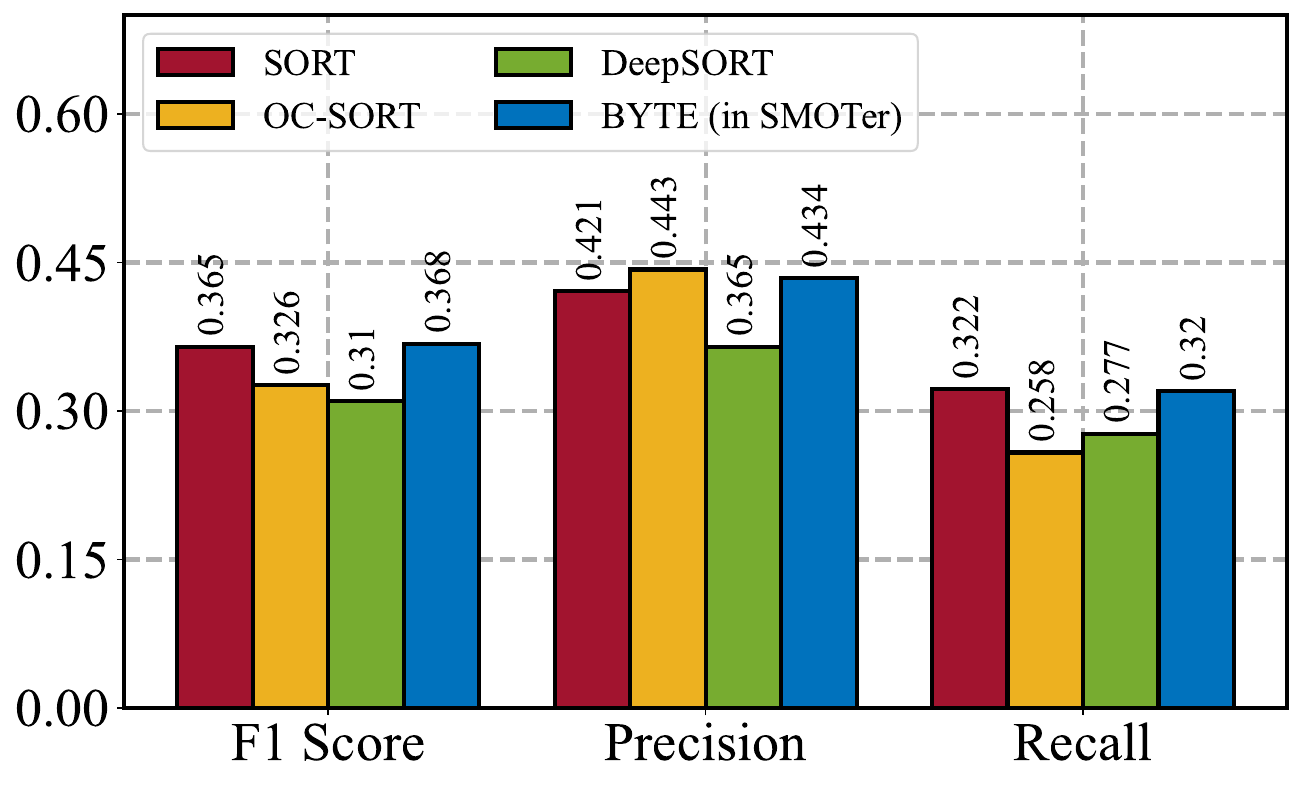}
        \captionsetup{labelformat=empty, font={scriptsize}, skip=2pt}
        \subcaption*{(d) Comparison results on interaction recognition}
    \end{minipage}
    
    \caption{Ablation experiments for different association strategies, including comparison results on object tracking in image (a), on video captioning in image (b), on instance captioning in image (c), and on interaction recognition in image (d).}
    \label{fig:ablation}
\end{figure*}

\subsection{Ablation Study}

\textbf{Impact of different feature fusion.}
Feature fusion is a crucial component of SMOTer for generating overall video and enhanced trajectory features. In this work, we study four types of feature fusion strategies for VFM and TFM, including attention-based fusion (using cross-attention), MLP-based fusion (using MLP module), concatenation-based fusion, and addition-based fusion. Please refer to \textbf{supplementary material} for detailed architectures. To assess the impact for VFM, we compare the performance on video caption generation because the fused video feature from VFM is used for this sub-task, while for TFM, we measure the results on instance captioning and interaction recognition since these two sub-tasks rely on the fused trajectory features. Tab.~\ref{tab:fusion-ablation} shows the results of different fusion mechanisms for VFM and TFM. We can see that attention-based fusion works generally better for VFM by achieving best ROUGE score of 0.261 and METEOR score of 0.223 and second best BLEU score of 0.245 and CIDEr score of 0.343 for video captioning. For TFM, concatenation-based fusion shows more superior results by obtaining best scores on ROUGE, METEOR, CIDEr for instance captioning and on Precision and F1 for interaction recognition.

\noindent
\textbf{Analysis on association mechanism.} Association is necessary in SMOTer to generate target trajectories. In our method, we leverage the popular BYTE~\cite{zhang2021fairmot} for proposal association. To analyze the impact of different association strategies, we compare other representative manners, including SORT~\cite{bewley2016simple}, DeepSORT~\cite{wojke2017simple}, and OC-SORT~\cite{cao2023observation}, with the adopted BYTE. Fig.~\ref{fig:ablation} demonstrates the comparison results of association mechanisms in SMOTer (using BYTE) and other approaches on four tasks, including object tracking (Fig.~\ref{fig:ablation} (a)), video captioning (Fig.~\ref{fig:ablation} (b)), instance captioning (Fig.~\ref{fig:ablation} (c)), and interaction recognition (Fig.~\ref{fig:ablation} (d)). From the results in Fig.~\ref{fig:ablation} (a)-(d), we can observe that BYTE used in our SMOTer can achieve the best or competitive performance on different metrics for various tasks, validating its effectiveness for semantic multi-object tracking.

\begin{table}[!t]\scriptsize
    \centering
    \renewcommand{\arraystretch}{1}
    \caption{Study of different thresholds $\tau_{p}$. The best result is highlighted in \textcolor{red}{red}.}
    \tabcolsep=0.8mm
    \resizebox{0.83\textwidth}{!}{
    \begin{tabular}{l|ccc|cc|cc|ccc}
        \toprule[1.2pt]
        & \multicolumn{3}{|c|}{\textbf{Tracking}} & \multicolumn{2}{|c|}{\textbf{Video Caption}} & \multicolumn{2}{|c|}{\textbf{Instance Caption}} & \multicolumn{3}{|c}{\textbf{Interaction}}\\
        \midrule[0.8pt]
        \textbf{$\tau_{p}$} & \textbf{HOTA}$\uparrow$ & \textbf{MOTA}$\uparrow$ & \textbf{IDF1}$\uparrow$ & \textbf{METEOR}$\uparrow$ & \textbf{CIDEr}$\uparrow$ & \textbf{METEOR}$\uparrow$ & \textbf{CIDEr}$\uparrow$ & \textbf{Prcn}$\uparrow$ & \textbf{Rcll}$\uparrow$ & \textbf{F1}$\uparrow$\\
        \midrule[0.8pt]
        0.1 & 70.51 & 72.30 & 78.80 & 0.219 & 0.256 & 0.192 & 0.069 & 0.378 & \textcolor{red}{0.378} & 0.319 \\
        0.2 & 71.10 & 73.80 & 78.90 & 0.215 & 0.289 & 0.199 & 0.069 & 0.384 & 0.365 & 0.331 \\
        0.3 & \textcolor{red}{71.98} & \textcolor{red}{77.71} & \textcolor{red}{80.65} & \textcolor{red}{0.224} & \textcolor{red}{0.343} & 0.209 & 0.087 & \textcolor{red}{0.434} & 0.320 & \textcolor{red}{0.368} \\
        0.4 & 71.19 & 76.36 & 76.01 & 0.214 & 0.330 & 0.215 & \textcolor{red}{0.091} & 0.420 & 0.302 & 0.341 \\
        0.5 & 68.30 & 74.34 & 78.90 & \textcolor{red}{0.224} & \textcolor{red}{0.343} & \textcolor{red}{0.216} & 0.087 & 0.396 & 0.291 & 0.310 \\
        \bottomrule[1.2pt]
    \end{tabular}
    \label{tab:throshold-ablation}
    }
\end{table}

\noindent
\textbf{Study of threshold $\tau_{p}$.} The parameter $\tau_{p}$ is used to remove low-confidence proposals to ensure the quality of subsequent trajectories, and thus crucial for all tasks in SMOTer. Because of this, we conduct an ablation study for investigating different $\tau_{p}$, and Tab.~\ref{tab:throshold-ablation} displays the results. From Tab.~\ref{tab:throshold-ablation}, we can observe that, when $\tau_{p}=0.3$, SMOTer achieves the overall best performance for different tasks.  

\subsection{Discussion}

\textbf{Semantic understanding for improving tracking.} SMOT aims to go beyond just tracking by incorporating additional semantic understanding tasks including trajectory-associated instance/video captioning and interaction recognition. We observe that these extra goals greatly improve the tracking performance. For example, as shown in Tab.~\ref{tab:comparison-tracking}, we can see that joint training of different tasks leads to better performance of SMOTer (71.98\%/77.71\% in HOTA/MOTA) in tracking compared to its baseline ByteTrack (68.84\%/73.87 in HOTA/MOTA). The possible reason is that SMOTer needs to comprehensively understand the video content for semantic multi-object tracking, and the learned semantic knowledge helps distinguish different object trajectories.

\noindent
\textbf{Challenge in SMOT.} One reason why SMOT is challenging lies in the complicated behaviors of targets, often leading to lengthy instance captions. For instance, a basketball player can be dribbling, shooting, and defending in a game. Models often miss some states in understanding. In BenSMOT, the average video length is about 23 seconds, with objects typically present for around 21 seconds, resulting in instance captions averaging over 35 words. This complexity poses a challenge for SMOT and indicates a potential area for future improvement.


\section{Conclusion}

In this paper, we introduce SMOT to expand the scope of MOT by integrating ``\emph{where}'' and ``\emph{what}''. To facilitate the research of SMOT, we propose the large-scale BenSMOT by including 3,292 videos from more than 40 diverse scenarios. To our best knowledge, BenSMOT is the first benchmark dedicated to SMOT. Furthermore, to encourage algorithm development on BenSMOT, we introduce SMOTer, an end-to-end tracker designed for  SMOT. Our results exhibit SMOTer surpasses offline-combination strategies, showing efficacy. By presenting BenSMOT and SMOTer, we hope to inspire more future research on SMOT.

\noindent
\textbf{Acknowledgements.} Heng Fan was not supported by any fund for this work.


%
%

\end{document}